\documentclass[accepted]{article}

\usepackage{newtxtext,newtxmath}

\usepackage{subcaption}

\usepackage{graphicx}

\usepackage{amsmath,bm}
\usepackage{textgreek}

\usepackage{natbib}

\usepackage{algorithm}
\usepackage{algorithmic}

\usepackage{hyperref}

\usepackage[per-mode=symbol]{siunitx}
\DeclareSIUnit\px{px}

\usepackage{soul}

\usepackage{hyperref}

\usepackage{amsopn}
\DeclareMathOperator{\EncD}{EncD}
\DeclareMathOperator{\DecD}{DecD}
\DeclareMathOperator{\Enc}{Enc}
\DeclareMathOperator{\Dec}{Dec}

\DeclareMathOperator{\logsoftmax}{LogSoftMax}
\DeclareMathOperator{\bce}{H}
\DeclareMathOperator{\mse}{MSE}

\newcommand{\getsplus}{\stackrel{+}\gets}

\usepackage{mathtools}
\DeclarePairedDelimiter\abs{\lvert}{\rvert}
\DeclarePairedDelimiter\norm{\lVert}{\rVert}

\newcommand{\beginsupplement}{%
        \setcounter{table}{0}
        \renewcommand{\thetable}{S\arabic{table}}%
        \setcounter{figure}{0}
        \renewcommand{\thefigure}{S\arabic{figure}}%
     }

\usepackage{icml2013}

\icmltitlerunning{Building the Integrated Cell}

\begin{document}

\twocolumn[
\icmltitle{Generative Modeling with Conditional Autoencoders:\\ Building an Integrated Cell}

\icmlauthor{Gregory R.\ Johnson}{gregj@alleninstitute.org}
\icmlauthor{Rory M.\ Donovan-Maiye}{rorydm@alleninstitute.org}
\icmlauthor{Mary M.\ Maleckar}{mollym@alleninstitute.org}

\icmladdress{Allen Institute for Cell Science,
            615 Westlake Ave N, Seattle, WA 98109}

\icmlkeywords{boring formatting information, machine learning, ICML}

\vskip 0.3in
]

\begin{abstract}
We present a conditional generative model to learn variation in cell and nuclear morphology and the location of subcellular structures from microscopy images.
Our model generalizes to a wide range of subcellular localization and allows for a probabilistic interpretation of cell and nuclear morphology and structure localization from fluorescence images.
We demonstrate the effectiveness of our approach by producing photo-realistic cell images using our generative model.
The conditional nature of the model provides the ability to predict the localization of unobserved structures given cell and nuclear morphology.
\end{abstract}

\section{Introduction}
\label{introduction}

A central biological principle is that cellular organization is strongly related to function.
Location proteomics \cite{Murphy:2005dd} addresses this by aiming to determine cell state -- i.e.\ subcellular organization -- by elucidating the localization of \emph{all} structures and how they change through the cell cycle, and in response to perturbations, e.g., mutation.
However, determining cellular organization is challenged by the multitude  of different molecular complexes and organelles that comprise living cells and drive their behaviors \cite{kim2014draft}.
Currently, the experimental state-of-the-art for live cell imaging is limited to the simultaneous visualization of only a limited number of tagged (2-6 tagged) molecules.
Modeling approaches can address this limitation by integrating subcellular structure data from diverse imaging experiments.
Due to the number and diversity of subcellular structures, it is necessary to build models that generalize well with respect to both representation and interpretation.

Image feature-based methods have previously been employed to describe and model cell organization \cite{boland2001neural, Carpenter:2006iu, Rajaram:2012di}.
While  useful for discriminative tasks, these approaches do not explicitly model the relationships between subcellular components, limiting the application to integration of all of these structures.

Generative models are useful in this context.
They capture variation in a population and encode it as a probability distribution, accounting for the relationships among structures.
Fundamental work has previously demonstrated the utility of expressing subcellular structure patterns as a generative model, which can then be used as a building block for models of cell behavior, i.e.\  \cite{Murphy:2005dd,donovan2016unbiased}.

Ongoing efforts to construct generative models of cell organization are primarily associated with the CellOrganizer project \cite{Zhao:2007is, Peng:2011et}.
That work implements a ``cytometric'' approach to modeling that considers the number of objects, lengths, sizes, etc.\ from segmented images and/or inverse procedural modeling, which can be particularly useful for both analyzing image content and approaching integrated cell organization.
These methods support parametric modeling of many subcellular structure types and, as such, generalize well when low amounts of appropriate imaging data are available.
However, these models may depend on preprocessing methods, such as segmentation, or other object identification tasks for which a ground truth is not available.
Additionally, there may exist subcellular structures for which a parametric model does not exist or may not be appropriate e.g.,  structures that vary widely in localization (diffuse proteins), or reorganize dramatically during e.g.\ mitosis or during a stimulated state (such as microtubules).

Thus, the presence of key structures for which current methods are not well suited motivates the need for a new approach that generalizes well to a wide range of structure localization.

Recent advances in adversarial networks \cite{Goodfellow:2014wp} are relevant to our problem.
They have the ability to learn distributions over images, generate photo-realistic exemplars, and learn sophisticated conditional relationships; see e.g.\ Generative Adversarial Networks \cite{Goodfellow:2014wp}, Varational Autoencoders/GAN \cite{Larsen:2015vi}, Adversarial Autoencoders \cite{Makhzani:2015tm}.

Leveraging these recent advances, we present a non-parametric model of cell shape and nuclear shape and location, and relate it to the variation of other subcellular components.
The model is trained on data sets of 300--750 fluorescence microscopy images; it accounts for the spatial relationships among these components, their fluorescent intensities, and generalizes well to a variety of localization patterns.
Using these relationships, the model allows us to predict the outcome of unobserved experiments, as well as encode complex image distributions into a low dimensional probabilistic representation.
This latent space serves as a compact coordinate system to explore variation.

In the following sections, we present the model, a discussion of the training and conditional modeling, and initial results which demonstrate its utility.
We then briefly discuss the results in context, current limitations of the work and future extensions.

\section{Model Description}
\label{description}

Our generative model serves several distinct but complementary purposes.
At its core, it is a probabilistic model of cell and nuclear shape (specifically, of cell shape and nuclear shape and \emph{location}) wedded to a probability distribution of structure localization (e.g.\ the localization of a certain protein) conditional on cell and nuclear shape.
This model, \emph{in toto}, can be used both as a classifier for images of localization pattern where the protein is unknown, and and as a tool with which one can predict the localization of unobserved structures \emph{de novo}.

The main components of our model are two autoencoders; one which encodes the variation in cell and nuclear shape, and another which learns the relationship between subcellular structures dependent on this encoding.

\begin{figure}[htbp]
\centering
\includegraphics[width=0.9\linewidth]{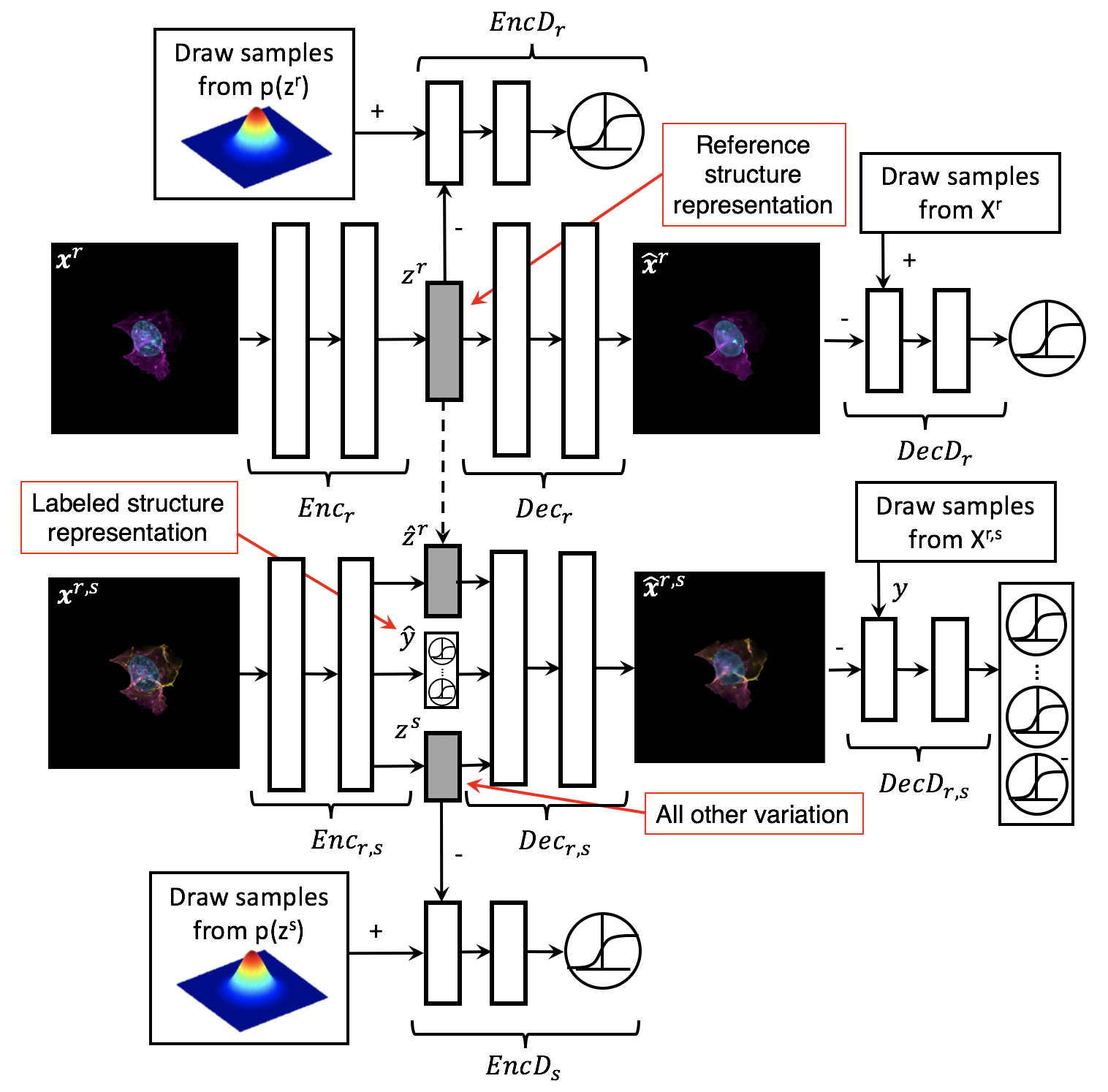}
\caption{The presented model.
The top half of the diagram outlines the reference structure model; the bottom half shows conditional model.
The parallel white boxes indicate a nonlinear function.
The model is a probabilistic model of cell and nuclear shape (specifically, of cell shape and nuclear shape and \emph{location}) wedded to a probability distribution of structure localization (e.g. the localization of a certain protein) conditional on cell and nuclear shape. This model can be used both as a classifier for images of localization pattern where the protein is unknown, and and as a tool for prediction of the localization of unobserved structures \emph{de novo}.
The main components are two autoencoders: one encoding the variation in cell and nuclear shape, and another which learns the relationship between subcellular structures dependent on this encoding. See Notation and Model description for details.
Figure adapted from \cite{Makhzani:2015tm}}
\label{fig:model}
\end{figure}

\subsection*{Notation}
The images input and output by the model are multi-channel (see figure~\ref{fig:cells_real}).
Each image $\bm{x}$ consists of both reference channels $r$ and a structure channel $s$.
Here, the cell and nuclear channels together serve as reference channels, and the structure channel varies, taking on one of the following structure types: \textalpha-actinin (actin bundles), \textalpha-tubulin (microtubules), \textbeta-actin (actin filaments), desmoplakin (desmosomes), fibrillarin (nucleolus), lamin B1 (nuclear membrane), myosin IIB (actomyosin bundles), Sec61\textbeta (endoplasmic reticulum), TOM20 (mitochondria), and ZO1 (tight junctions).
We denote which content is being used by the use of superscripts; $\bm{x}^{r,s}$ indicates all channels are being used, whereas $\bm{x}^{s}$ indicates only the structure channel is being used, and $\bm{x}^{r}$ only the reference channels.
We use $y$ to denotes an index-valued categorical variable indicating which structure type is labeled in $\bm{x}^{s}$.
For example, $y=1$ might correspond to the \textalpha-actinin channel being active, $y=2$ to the \textalpha-tubulin channel, etc.
While $y$ is a scalar integer, we also use $\bm{y}$, a one-hot vector representation of $y$, with a one in the $y$th element of $\bm{y}$ and zeros elsewhere.

\subsection{Model of cell and nuclear variation} \label{subsub:model_cellnuc}

We model cell shape and nuclear shape using an autoencoder to construct a latent-space representation of these reference channels.
The model (figure~\ref{fig:model}, upper half) attempts to map images of reference channels to a multivariate normal distribution of moderate dimension -- here we use a sixteen dimensional distribution.
The choice of a normal distribution as the prior for the latent space is in many respects one of convenience, and of small consequence to the model.
The nonlinear mappings learned by the encoder and decoder are coupled to both the shape and dimensionality of the latent space distribution; the mapping and the distribution only function in tandem -- see e.g.\ figure~4 in \cite{Makhzani:2015tm}.

The primary architecture of the model is that of an autoencoder, which itself consists of two networks: an encoder $\Enc_r$ that maps an image $\bm{x}$ to a latent space representation $\bm{z}$ via a learned deterministic function $q(\bm{z}^r|\bm{x}^r)$, and a decoder $\Dec_r$ to reconstruct samples from the latent space representation using a similarly learned function $g(\hat{\bm{x}}^r|\bm{z}^r)$.

We use the following notation for these mappings:
\begin{align}
    \bm{z}^r &= q(\bm{z}^r|\bm{x}^r) = \Enc_r(\bm{x}^r) \\
    \hat{\bm{x}}^r &= g(\hat{\bm{x}}^r|\bm{z}^r) = \Dec_r(\bm{z}^r)
\end{align}
where an input image $\bm{x}$ is distinguished from a reconstructed image $\hat{\bm{x}}$ by the hat over the vector.

\subsubsection{Encoder and Decoder}

The autoencoder minimizes the pixel-wise binary cross-entropy loss of the input and reconstructed input using binary cross entropy,
\begin{align} \label{eqn:loss_xr}
    \mathcal{L}_{x^r} &= \bce( \hat{\bm{x}}^r, \bm{x}^r)
\end{align}
where
\begin{align}
    \bce(\bm{\hat{u}}, \bm{u}) =
        -\tfrac{1}{n} \sum_{p}
            u_p \log{\hat{u}_p} +
            (1 - u_p) \log{(1-\hat{u}_p)}
\end{align}
and the sum is over all the pixels $p$ in all the channels in the images $\bm{u}$.
We use this function for all images regardless of content (i.e.\ we use it for $\bm{x}^r$ and $\bm{x}^{r,s}$)

\subsubsection{Encoding Discriminator}

In addition to minimizing the above loss function, the autoencoder's latent space -- the output of $\Enc_r$ -- is regularized by the use of a discriminator $\EncD_r$, the encoding discriminator.
This discriminator $\EncD_r$ attempts to distinguish between latent space embeddings that are mapped from the input data, and latent space embeddings that are generative drawn from the desired prior latent space distribution (which here is a sixteen dimensional multivariate normal).
In attempting to fool the discriminator, the autoencoder is forced to learn a latent space distribution $q(\bm{z}^r)$ that is similar in form to the prior distribution $p(\bm{z}^r)$ \cite{Makhzani:2015tm}.

The encoding discriminator $\EncD_r$ is trained on samples from both the embedding space $\bm{z} \sim q(\bm{z}^r)$ and from the desired prior $\tilde{\bm{z}} \sim p(\bm{z}^r)$.
We refer to $\bm{z}$ as observed samples, and $\tilde{\bm{z}}$ as generated samples, and use the subscripts obs and gen to indicate these labels.
Trained on these samples, $\EncD_r$ outputs a continuous estimate of the source distribution, $\hat{v}^{\EncD_r} \in (0,1)$.

The objective function for the encoding discriminator is thus to minimize the binary-cross entropy between the true labels $v$ and the estimated labels $\hat{v}$ for generated and observed images:
\begin{align} \label{eqn:loss_EncDr}
    \mathcal{L}_{\EncD_r} &=    \bce(\hat{v}^{z^r}_{\text{gen}}, v^{z^r}_{\text{gen}}) +
                                \bce(\hat{v}^{z^r}_{\text{obs}}, v^{z^r}_{\text{obs}})
\end{align}
%

\subsubsection{Decoding Discriminator}

The final component of the autoencoder for cell and nuclear shape is an additional adversarial network $\DecD_r$, the decoding discriminator, which operates on the output of the decoder to ensure that the decoded images are representative of the data distribution, similar to that of \cite{Larsen:2015vi}.
We train $\DecD_r$ on images from the data distribution, $\bm{x}^r_\text{obs} \sim \bm{X}^r$, which we refer to as observed images, and on decoded draws from the latent space, $\bm{x}^r_\text{gen} \sim \Dec_r(\tilde{\bm{z}}^r)$, which we refer to as generated images.
The loss function for the decoding discriminator is then:
\begin{align}
    \mathcal{L}_{\DecD_r} &=    \bce(\hat{v}^{x^r}_{\text{gen}}, v^{x^r}_{\text{gen}}) +
                                \bce(\hat{v}^{x^r}_{\text{obs}}, v^{x^r}_{\text{obs}})
\end{align}

\subsection{Conditional model of structure localization}

Given a trained model of cell and nuclear shape variation from the above network component, we then train a conditional model of structure localization localization upon the learned cell and nuclear shape model.
This model (figure~\ref{fig:model}, lower half) consists of several parts, similar to those above: the core is a tandem encoder $\Enc_{r,s}$ and decoder $\Dec_{r,s}$ that encode and decode images to and from a low dimensional latent space; in addition, a discriminative decoder $\EncD_{s}$ regularizes the latent space, and a discriminative decoder $\DecD_{r,s}$ ensures that the decoded images are similar to the input distribution.

\subsubsection{Conditional Encoder}

The encoder $\Enc_{r,s}$ is given images containing both the reference structure and structures of protein localization, $\bm{x}_{r,s}$ and produces three outputs:
\begin{align} \label{eqn:cond_enc}
    {\hat{\bm{z}}^r, \hat{\bm{y}}, \bm{z}^s} &= \Enc_{r,s}(\bm{x}^{r,s}) = q({\hat{\bm{z}}^r, \hat{\bm{y}}, \bm{z}^s} | \bm{x}^{r,s})
\end{align}
Here $\hat{\bm{z}}^r$ is the reconstructed cell and nuclear shape latent-space representation learned in Section~\ref{subsub:model_cellnuc}, $\hat{\bm{y}}$ is an estimate of which structure channel was learned, and $\bm{z}^s$ is a latent variable that encodes all remaining variation in image content not due to cell/nuclear shape and structure channel.
Therefore $\bm{z}^s$ is learned dependent on the latent space embeddings of the reference structure, $\bm{z}^r$.

The loss function for the reconstruction of the latent space embedding of the cell and nuclear shape is the mean squared error between the embedding $\bm{z}^r$ learned from the cell and nuclear shape autoencoder and the estimate $\hat{\bm{z}}^r$ of that embedding produced by the conditional portion of the model:
\begin{align}
     \mathcal{L}_{\hat{\bm{z}}^r} = \mse (\bm{z}^r, \hat{\bm{z}}^r) =
                                    \tfrac{1}{n} \norm{ \bm{z}^r - \hat{\bm{z}}^r}^2
\end{align}

The output $\hat{\bm{y}}$ in equation~\ref{eqn:cond_enc} is a probability distribution over structure channels, giving an estimate of the class label for the structure.
In our notation, $y$ is an integer value representing the true structure channel, and takes an integer value $1 \ldots K$, while $\bm{y}$ is the one-hot encoding of that label, a vector of length $K$ equal to 1 at the $y$th position and 0 otherwise.
Similarly, $\hat{y}$ is a vector of length $K$ whose $k$th element represents the probability of assigning the label $y = k$.

We use the softmax function to assign these probabilities.
In general, the softmax function is given by
\begin{align}
    \logsoftmax(\bm{u}, i) &= \log \left(\frac{e^{\bm{u}_i}}{\sum_j e^{\bm{u}_j}} \right)
\end{align}
the loss function for $\hat{\bm{y}}$ is then
\begin{align}
    \mathcal{L}_{y} = -\logsoftmax \left( \hat{\bm{y}}, y \right)
\end{align}

The final output of the conditional encoder $\bm{z}^s$ can be interpreted as a variable that encodes the variation in the localization of the labeled structure independent of cell and nuclear shape.

\subsubsection{Encoding Discriminator}

The latent variable $\bm{z}^s$ is similarly regularized by an adversary $\EncD_s$ that enforces the distribution of this latent variable be similar to a chosen prior $p(\bm{z}^s)$.
The loss function for the adversary takes the same form as equation~\ref{eqn:loss_EncDr}:
\begin{align} \label{eqn:loss_EncDs}
    \mathcal{L}_{\EncD_r} &=    \bce(\hat{v}^{z^s}_{\text{gen}}, v^{z^s}_{\text{gen}}) +
                                \bce(\hat{v}^{z^s}_{\text{obs}}, v^{z^s}_{\text{obs}})
\end{align}
%

\subsubsection{Conditional Decoder}

The conditional decoder $\Dec_{r,s}$ outputs the image reconstruction given the latent space embedding $\hat{\bm{z}}^r$, the class estimator $\hat{\bm{y}}$, and the structure channel variation~$\bm{z}^s$:
\begin{align}
    \hat{\bm{x}}^{r,s} &= \Dec_{r,s}({\hat{\bm{z}}^r, \hat{\bm{y}}, \bm{z}^s}) = g(\bm{x}^r | {\hat{\bm{z}}^r, \hat{\bm{y}}, \bm{z}^s})
\end{align}

The loss function for image reconstruction takes  the same form as equation~\ref{eqn:loss_xr}, the binary cross entropy between the input and reconstructed image:
\begin{align}
    \mathcal{L}_{\bm{x}^{r,s}} &= \bce( \hat{\bm{x}}^{r,s}, \bm{x}^{r,s}).
\end{align}

\subsubsection{Decoding Discriminator}

As in the cell and nuclear shape model, attached to the decoder $\Dec_{r,s}$ is an adversary $\DecD_{r,s}$ intended to enforce that the reconstructed images are similar in distribution to the input images.
The output of this discriminator is a vector $\hat{\bm{y}}^{\DecD_{r,s}}$ that has $\abs{\bm{y}}+1 = K+1$ output labels, which take a value in $[1, \ldots , K, \text{gen}]$.
That is, $\hat{\bm{y}}^{\DecD_{r,s}}$ has one slot for real images of each particular labeled structure channel, and one additional slot for reconstructed (aka, generated) images of all channels.
The loss function is therefore
\begin{align}
    \mathcal{L}_{\DecD_{r,s}} = -\logsoftmax \big( \hat{\bm{y}}^{\DecD_{r,s}}, y \big)
\end{align}

\subsection{Training procedure}

The training procedure occurs in two phases.
We first train the model of cell and nuclear shape variation, components $\Enc^r$, $\Dec^r$, $\EncD^r$, $\DecD^r$, to convergence (algorithm~\ref{alg:train_reference}).
We then train the conditional model, components $\Enc^{r,s}$, $\Dec^{r,s}$, $\EncD^s$, $\DecD^{r,s}$ (algorithm~\ref{alg:train_conditional}).

In training the model, we adopt three strategies from \cite{Larsen:2015vi}: we limit error signals to relevant networks by propagating the gradient update from any $\DecD$ through only $\Dec$, we update decoders with respect Adversarial discrimination of generated and reconstructed images, and we weight the gradient update from the discriminators with the scalars $\gamma_{\Enc}$ and $\gamma_{\Dec}$.
The parameters are therefore updated as follows:
\begin{align}
    \theta_{\Enc_{r}} &\getsplus \nabla_{\theta_{\Enc_{r}}} ( \mathcal{L}_{x_r} + \gamma_{\Enc}\mathcal{L}_{\EncD_s}) \\
    \theta_{\Dec_{r}} &\getsplus \nabla_{\theta_{\Dec_{r}}} ( \mathcal{L}_{x_r} + \gamma_{\Dec}\mathcal{L}_{\DecD_{s}} ) \\
    \theta_{\Enc_{r,s}} &\getsplus \nabla_{\theta_{\Enc_{r,s}}} ( \mathcal{L}_{x_{r,s}} + \mathcal{L}_{\hat{z}_r} + \mathcal{L}_{y} + \gamma_{\Enc}\mathcal{L}_{\EncD_s}) \\
    \theta_{\Dec_{r,s}} &\getsplus \nabla_{\theta_{\Dec_{r,s}}} ( \mathcal{L}_{x_{r,s}} + \gamma_{\Dec}\mathcal{L}_{\DecD_{r,s}} )
\end{align}

\begin{algorithm}[htbp]
   \caption{Training procedure reference structure model}
   \label{alg:train_reference}
\begin{algorithmic}
   \STATE $\theta_{\Enc_{r}}$, $\theta_{\Dec_{r}}$, $\theta_{\EncD_{r}}$, $\theta_{\DecD_{r}}$ $\gets$ initialize network parameters
   \REPEAT
   \STATE $\bm{X^r} \gets$ random mini-batch from reference set
   \STATE $\bm{Z}^r \gets \Enc_{s}(\bm{X}^{r})$
   \STATE $\bm{\hat{X}^{r}} \gets \Dec_{r}(\hat{\bm{Z}}^{r})$

   \STATE $\hat{\bm{V}}^{\EncD_{r}}_\text{gen} \gets \EncD_{r}(\bm{\tilde{Z}}^r)$
   \STATE $\hat{\bm{V}}^{\EncD_{r}}_\text{obs} \gets \EncD_{r}(\bm{Z}^r)$

   \STATE $\hat{\bm{V}}^{\DecD_{r}}_\text{obs} \gets \DecD_{r}(\bm{X}^{r})$
   \STATE $\hat{\bm{V}}^{\DecD_{r}}_\text{gen} \gets \DecD_{r}(\Dec(\tilde{\bm{Z}}^r))$

   \STATE $\mathcal{L}_{\DecD_{r}} \gets \bce(\hat{\bm{V}}^{\DecD_{r}}_\text{obs}, \bm{V}_{\text{obs}})$ \par
   \hskip\algorithmicindent $ + \bce(\hat{\bm{V}}^{\DecD_{r}}_\text{gen}, \bm{V}_{\text{gen}})$

   \STATE $\theta_{\DecD_{r}} \getsplus \nabla_{\theta_{\DecD_{r}}} \mathcal{L}_{\DecD_{r}} $

   \STATE $\mathcal{L}_{\EncD_{r}} \gets \bce(\hat{\bm{V}}^{\EncD_{r}}_\text{gen}, \bm{V}_{\text{gen}})$ \par
   \hskip\algorithmicindent $ + \bce(\hat{\bm{V}}^{\EncD_{r}}_\text{obs}, \bm{V}_{\text{obs}})$
   \STATE $\theta_{\EncD_{r}} \getsplus \nabla_{\theta_{\EncD_{r}}} \mathcal{L}_{\EncD_{r}}$

   \STATE $\mathcal{L}_{\hat{X}^r} \gets \bce(\hat{\bm{X}}^r, \bm{X}^r)$

   \STATE $\mathcal{L}_{\EncD_r} \gets \bce(\hat{\bm{V}}^{\EncD_{r}}_\text{obs}, \bm{V}_{\text{gen}})$
   \STATE $\mathcal{L}_{\DecD_r} \gets \bce(\hat{\bm{V}}^{\DecD_{r}}_\text{gen}, \bm{V}_{\text{obs}}) + \bce(\DecD_r(\hat{\bm{X}}^r), \bm{V}_{\text{obs}})$

   \STATE $\theta_{\Enc_{r}} \getsplus \nabla_{\theta_{\Enc_{r}}} \mathcal{L}_{\hat{X}^r} + \gamma_{\Enc}\mathcal{L}_{\EncD_{r}}$
   \STATE $\theta_{\Dec_{r}} \getsplus \nabla_{\theta_{\Dec_{r}}} \mathcal{L}_{\hat{X}^r} + \gamma_{\Dec}\mathcal{L}_{\DecD_{r}}$

   \UNTIL{convergence}

\end{algorithmic}
\end{algorithm}

\begin{algorithm}[htbp]
   \caption{Training procedure for conditional relationship model}
   \label{alg:train_conditional}
\begin{algorithmic}

   \STATE $\theta_{\Enc_{r,s}}$, $\theta_{\Dec_{r,s}}$, $\theta_{\EncD_{s}}$, $\theta_{\DecD_{r,s}}$ $\gets$ initialize network \par
   \hskip\algorithmicindent parameters

   \REPEAT
   \STATE $\bm{X}^{r,s}, Y, \bm{Z}^r \gets$ random mini-batch \par
   \hskip\algorithmicindent from reference and structure set
   \STATE $\hat{\bm{Z}}^r, \hat{\bm{Y}}, \bm{Z}^s \gets \Enc_{r,s}(\bm{X}^{r,s})$
   \STATE $\bm{\hat{X}}^{r,s} \gets \Dec_{s}(\hat{\bm{Z}}^{r}, \hat{\bm{Y}}, \bm{Z}^{s})$

   \STATE $\hat{\bm{V}}^{\EncD_{s}}_\text{gen} \gets \EncD_{s}(\bm{\tilde{Z}}^s)$
   \STATE $\hat{\bm{V}}^{\EncD_{s}}_\text{obs} \gets \EncD_{s}(\bm{Z}^s)$


   \STATE $\hat{\bm{Y}}_{\text{obs}} \gets \DecD_{r,s}(\bm{X}^{r,s})$
   \STATE $\hat{\bm{Y}}_{\text{gen}} \gets \DecD_{r,s}(\Dec(\hat{\bm{Z}}^r, \hat{\bm{Y}}, \tilde{\bm{Z}}^{s}))$

   \STATE $\mathcal{L}_{\EncD_{s}} \gets \bce(\hat{\bm{V}}^{\EncD_{r}}_\text{gen}, \bm{V}_{\text{gen}}) + \bce(\hat{\bm{V}}^{\EncD_{s}}_\text{obs}, \bm{V}_{\text{obs}})$
   \STATE $\theta_{\EncD_{s}} \getsplus \nabla_{\theta_{\EncD_{s}}} \mathcal{L}_{\EncD_{s}}$

   \STATE $\mathcal{L}_{\DecD_{r,s}} \gets -\logsoftmax \big(\hat{\bm{Y}}_{\text{obs}}, Y \big)$ \par
   \hskip\algorithmicindent $ - \logsoftmax \big(\hat{\bm{Y}}_{\text{gen}}, Y_{\text{gen}} \big)$
   \STATE $\theta_{\DecD_{r,s}} \getsplus \nabla_{\theta_{\DecD_{r,s}}}  \mathcal{L}_{\DecD_{r,s}}$

   \STATE $\mathcal{L}_{\hat{X}^{r,s}} \gets \bce(\hat{\bm{X}}^{r,s}, \bm{X}^{r,s})$
   \STATE $\mathcal{L}_{Y} \gets -\logsoftmax(\hat{\bm{Y}}, Y)$
   \STATE $\mathcal{L}_{\hat{\bm{Z}}^r} \gets \mse(\hat{\bm{Z}}^r, \bm{Z}^r)$

   \STATE $\mathcal{L}_{\EncD_{s}} \gets \bce(\hat{\bm{V}}^{\EncD_{s}}_\text{obs}, \bm{V}_{\text{gen}})$
   \STATE $\mathcal{L}_{\DecD_{r,s}} \gets -\logsoftmax(\hat{\bm{Y}}_{\text{gen}}, Y) $ \par
   \hskip\algorithmicindent $ -\logsoftmax(\DecD_{r,s}(\hat{\bm{X}}^{r,s}), Y)$

   \STATE $\theta_{\Enc_{r,s}} \getsplus \nabla_{\theta_{\Enc_{r,s}}} \mathcal{L}_{\hat{X}^{r,s}} + \mathcal{L}_{Y} + \mathcal{L}_{\hat{\bm{Z}}^r} + \gamma_{\Enc}\mathcal{L}_{\EncD_{s}}$
   \STATE $\theta_{\Dec_{r,s}} \getsplus \nabla_{\theta_{\Dec_{r,s}}} \mathcal{L}_{\hat{X}^{r,s}} + \gamma_{\Dec}\mathcal{L}_{\DecD_{r,s}}$

   \UNTIL{convergence}
\end{algorithmic}
\end{algorithm}

\subsection{Integrative Modelling}
Beyond encoding and decoding images, we are able to leverage the conditional model of structure localization given cell and nuclear shape as a tool to predict the localization of unobserved structures, $p(x^s|x^r, y)$.
In particular, we use the maximum likelihood  structure localization given the cell and nuclear channels.
The procedure for predicting this localization is shown in algorithm~\ref{alg:integration}.

\begin{algorithm}[htbp]
   \caption{Structure integration procedure}
   \label{alg:integration}
\begin{algorithmic}
   \STATE trained $\Enc_{r}$ and $\Dec_{r,s}$
   \STATE $\bm{x}^r \gets$ reference structure image
   \STATE $\bm{z}^r \gets \Enc_{r}(\bm{x}^{r})$
   \FOR{\textbf{each} structure in structures}
   \STATE $\bm{y} \gets$ structure
   \STATE $\bm{z}^s \gets \text{argmax}_{z_{s}} p(z_{s})$ 
   \STATE $\bm{\hat{x}}_{r,s} \gets \Dec_{r,s}(\bm{z}^r, \bm{y}, \bm{z}^s)$
   \STATE append $\hat{\bm{x}}_{s}$ to $\bm{x}_{\text{out}}$
   \ENDFOR
\end{algorithmic}
\end{algorithm}

\section{Results}
\label{results}

\subsection{Data Set}

\begin{figure}[htbp]
\centering
\includegraphics[width=0.9\linewidth]{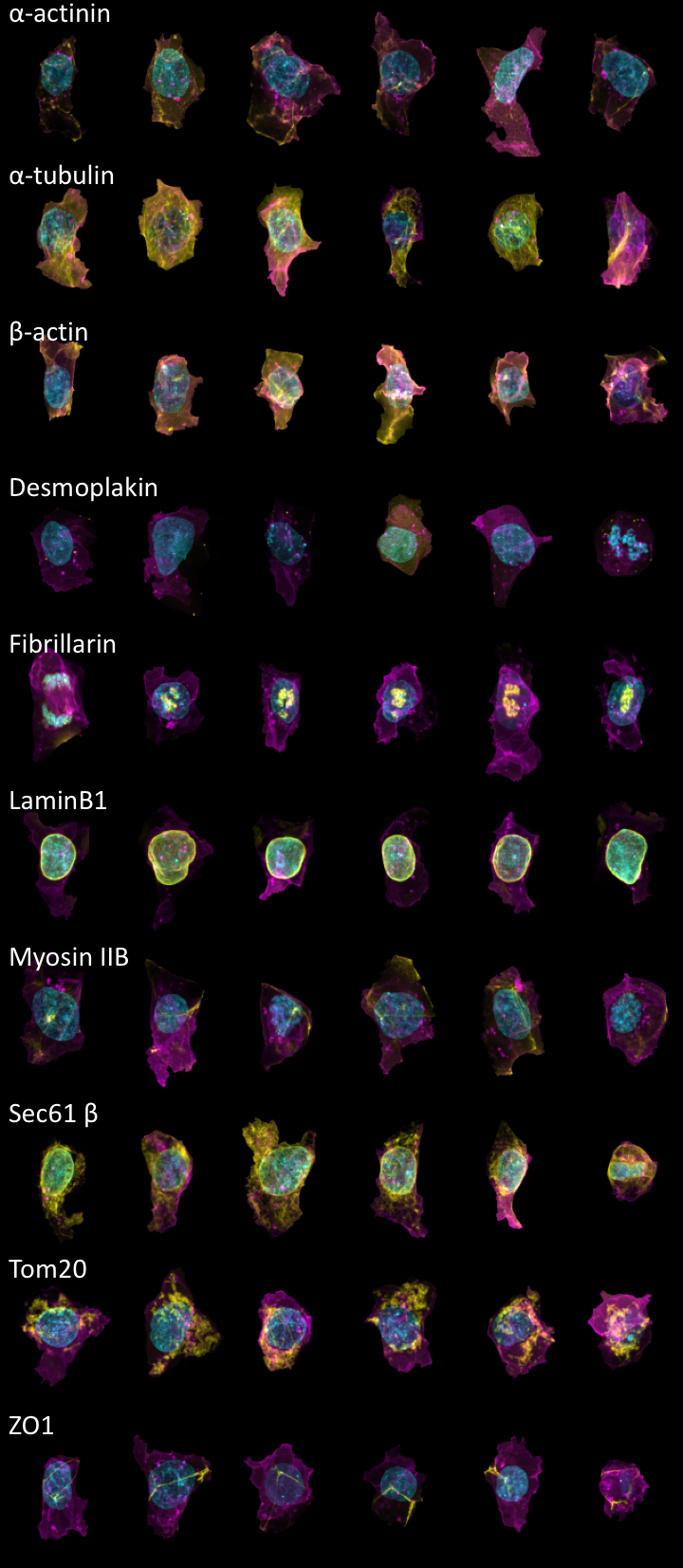}
\caption{Example images for each of the 10 labeled structures of focus in this paper.
Rows correspond to observed microscopy images, used as inputs to the model, for six arbitrary cells, each with a particular fluorescently labeled structure as named, shown in yellow.
The reference structures, the cell membrane and nucleus (DNA), are shown in magenta and cyan, respectively.
Images have been cropped for visualization purposes.
See figure \ref{fig:cells_real_struct_only} for isolated observed structure channel only.}
\label{fig:cells_real}
\end{figure}

For the experiments presented here, we use a collection of 2D segmented cell images generated from a maximum intensity projection of a 3D confocal microscopy data set from human induced pluripotent stem cells gene edited to express mEGFP on proteins that localize to specific structures, e.g.\ \textalpha-actinin (actin bundles), \textalpha-tubulin (microtubules), \textbeta-actin (actin filaments), desmoplakin (desmosomes), fibrillarin (nucleolus), lamin B1 (nuclear membrane), myosin IIB (actomyosin bundles), Sec61\textbeta (endoplasmic reticulum), TOM20 (mitochondria), and ZO1 (tight junctions).
Details of the source image collection are available via the Allen Cell Explorer at \url{http://allencell.org}.
Briefly, each image consists of channels corresponding to the nuclear signal, cell membrane signal, and a labeled sub-cellular structure of interest (see figure~\ref{fig:cells_real}).
Individual cells were  segmented, and each channel was processed by subtracting the most populous pixel intensity, zeroing-out negative-valued pixels, rescaling image intensity between 0 and 1, and max-projecting the 3D image along the height-dimension.
The cells were aligned by the major axis of the cell shape, and centered according to the center of mass of the segmented nuclear region, and flipped according to image skew.
Each of the 6077 cell images were rescaled to \SI{0.317}{\micro\metre\per\px}, and padded to $256 \times 256$ pixels.
The model took approximately 16 hours to train on one Pascal Titan X GPU.

\subsection{Model implementation}

A summary of the model architectures is described in Section~\ref{sec:architecture}.
We based the architectures and their implementations on a combination of resources, primarily \cite{Larsen:2015vi, Makhzani:2015tm, Radford:2015wf}, and Kai Arulkumaran's Autoencoders package \cite{Arulkumaran:2017}.

We found that adding white noise to the first layer of decoder adversaries, $\DecD^r$ and $\DecD^{r,s}$, stabilizes the relationship between the adversary and the autoencoder and improves convergence as in \cite{Sonderby:2016ta} and \cite{Salimans:2016wg}.

We choose a sixteen dimensional latent space for both  $Z^r$ and $Z^s$.

\subsection{Training}

To train the model, we used the Adam optimizer \cite{Kingma:2014us} to perform gradient-descent, with a batch size of 32, learning rate of 0.0002 for all model components ($\Enc_r$, $\Dec_r$, $\EncD_r$, $\DecD_r$, $\Enc_{r,s}$, $\Dec_{r,s}$, $\EncD_s$, $\DecD_{r,s}$), with $\gamma_{\Enc}$ and $\gamma_{\Dec}$ values of $10^{-4}$ and $10^{-5}$ respectively.
The dimensionality of the latent spaces $\bm{Z}^r$ and $\bm{Z}^s$ were set to 16, and the prior distribution for both is an isotropic gaussian.

We spit the data set into 95\% training and 5\% test (for more details see table~\ref{tbl:traintest}), and trained the model of cell and nuclear shape for 150 epochs, and the conditional model for 220 epochs.
The model was implemented in Torch7 \cite{Collobert_torch7:a}, and ran on an Nvidia Pascal TitanX.
The model took approximately 16 hours to train.
Further details of our implementation can be found in the software repository.

The training curves for the reference and conditional model are shown in figure~\ref{fig:traincurves}.

\subsection{Experiments}

We performed a variety of ``experiments'' exploring the utility of our model architecture.
While quantitative assessment is paramount, the nature of the data makes qualitative assessment indispensable as well, and we include experiments of this type in addition to more traditional  measures of performance.

\subsubsection{Image reconstruction}

A necessary but not sufficient condition for our model to be of use is that the images of cells reconstructed from their latent space representations bear some semblance to the native images.
Examples of image reconstruction from the training and test set are shown in figure~\ref{fig:traintest_ref} for our reference structures and figure~\ref{fig:traintest_struct} for the structure localization model.
As seen in the figures, the model is able to recapitulate the essential localization patterns in the cells, and produce accurate reconstructions in both the training and test data.

\subsubsection{Latent space representation}

We explored the generative capacity of our model by mapping out the variation in cell morphology due to traversal of the latent space.
Since the latent spaces in our model are sixteen dimensional and isotropic, dimensionality reduction techniques are of little value, and we resorted to mapping 2D slices of the space.

To demonstrate this variation is smooth, we plot the first two dimensions of the latent space for cell and nuclear shape variation are shown in figure~\ref{fig:embed_ref}.
The first two dimensions of the latent space for structure variation are shown in figure~\ref{fig:embed_struct}.
In both figures, the orthogonal dimensions are set to their MLE value of zero.

\subsubsection{Image Classification}

While classification is not our primary use-case, it is a worthwhile benchmark of a well-functioning multi-class generative model.
To evaluate the performance of the class-label identification of $\Enc^{r,s}$ we compared the results of the predicted labels and true labels on our hold out set.
A summary of the results of our multinomial classification task is shown in table~\ref{tbl:confmat}.
As seen in the table, our model is able to accurately classify most structure, and has trouble only on the poorly sampled or underrepresented classes.

\subsubsection{Integrating Cell Images}

\begin{figure}[htbp]
\centering
\includegraphics[width=0.9\linewidth]{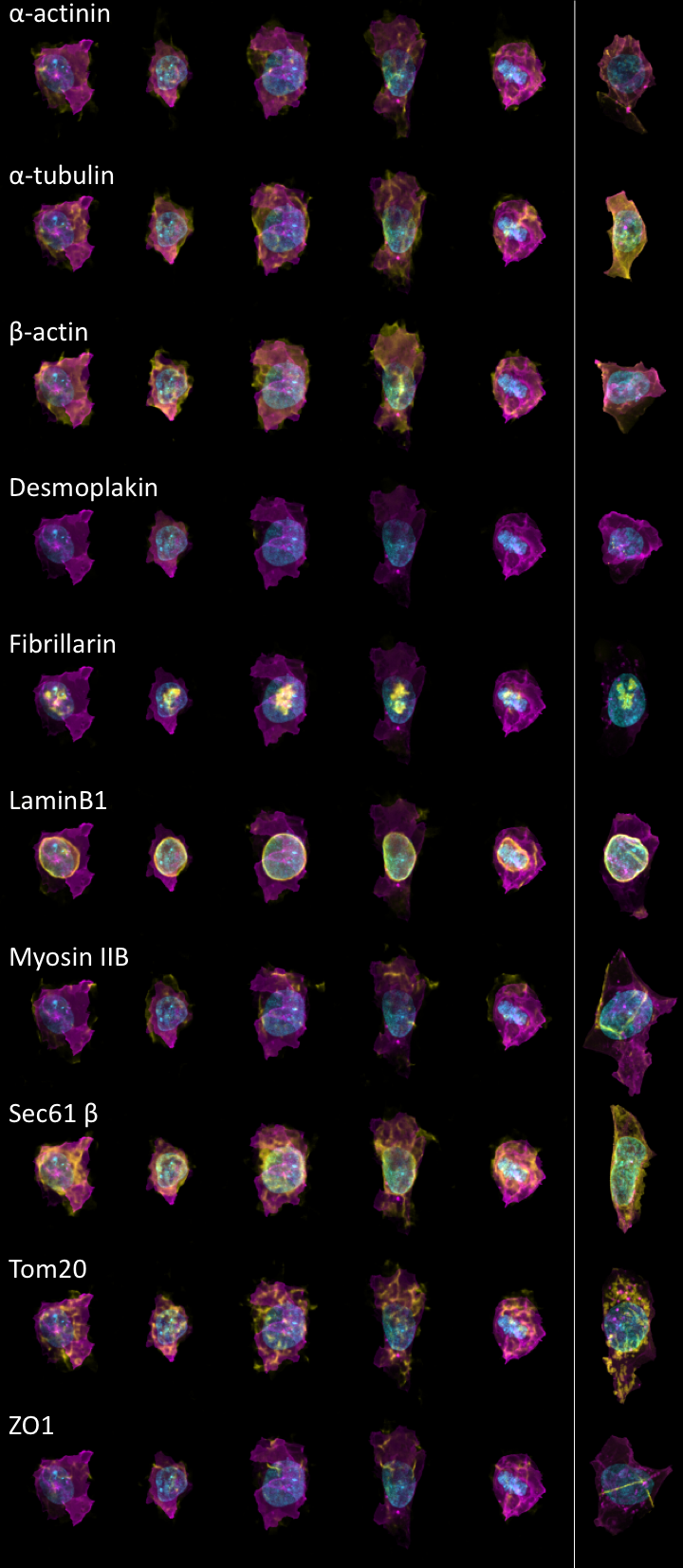}
\caption{Most probable localization patterns predicted for selected cells for each structure (rows, top to bottom, structure as labeled, shown in yellow).
The first 5 columns show the maximum likelihood of localization for each structure, given the cell and nuclear shape.
The last column (far right) shows an experimentally observed cell with that labeled structure for comparison.
As before, reference structures, cell membrane and nucleus (DNA), are in magenta and cyan, respectively.
Images have been cropped for visualization purposes.
Note for example how fibrillarin resides within the DNA, and lamin B1 surrounds the DNA.
See figure \ref{fig:pred_struct_struct_only} for structure channel only.}
\label{fig:pred_struct}
\end{figure}

Conditional upon the cell and nuclear shape, we predict the most likely position of any particular structure via algorithm~\ref{alg:integration}.
Some examples of the maximum likelihood estimate of structure localization given cell and nuclear shapes is shown in figure~\ref{fig:pred_struct}.

\section{Discussion}
\label{discussion}

Building models that capture relationships between the morphology and organization of cell structures is a difficult problem.
While previous research has focused on constructing application-specific parametric approaches, due to the the extreme variation in localization among difference structures, these approaches may not be convenient to employ for all structures under all conditions.
Here, we have presented a nonparametric conditional model of structure organization that generalizes well to a wide variety of localization patterns, encodes the variation in cell structure and organization, allows for a probabilistic interpretation of the image distribution, and generates high quality synthetic images.

Our model of cell and subcellular structure differs from previous generative models \cite{Zhao:2007is,Peng:2011et,Johnson:2015bz}: we directly model the localization of fluorescent labels, rather than the detected objects and their boundaries.
While object segmentation can be essential in certain contexts, and helpful in others, when these approaches are not necessary, it can be advantageous to omit these non-trivial intermediate steps.
Our model does not constitute a ``cytometric'' approach (i.e.\ counting objects), but due to the fact that we are directly modeling the localization of signal, we drastically reduce the modeling time by minimizing the amount of segmentation and the task of evaluating this segmentation with respect to the ``ground truth''.

Even considering these these differences, our model is compatible with existing frameworks and will allow for mixed parametric and non-parametric localization relationships, where our model can be used for predicting localization of structures when an appropriate parametric representation may not exist.

Our model permits several straightforward extensions, including the obvious extension to modeling cells in three dimensions.
Because of the flexibility of our latent-space representation, we can potentially encode information such as position in the cell cycle, or along a differentiation pathway.
Given sufficient information, it would be possible to encode a representation of ``structure space'' to predict the localization of unobserved structures, or ``perturbation space'', such as in \cite{Paolini:2006be}, and potentially couple this with active learning approaches \cite{Naik:2016gw} to build models that learn and encode the localization of diverse subcellular structures under different conditions.

\section*{Software and Data}

The code for running the models used in this work is available at \url{https://github.com/AllenCellModeling/torch_integrated_cell}

The data used to train the model is available at \url{s3://aics.integrated.cell.arxiv.paper.data}.

\section*{Acknowledgements}

We would like to thank Robert F.\ Murphy, Julie Theriot, Rick Horwitz, Graham Johnson, Forrest Collman, Sharmishtaa Seshamani and Fuhui Long for their helpful comments, suggestions, and support in the preparation of the manuscript.

Furthermore, we would like to thank all members of the Allen Institute for Cell Science team, who generated and characterized the gene-edited cell lines, developed image-based assays, and recorded the high replicate data sets suitable for modeling. We particularly thank Liya Ding for segmentation data. These contributions were absolutely critical for model development.

We would like to thank Paul G.\ Allen, founder of the Allen Institute for Cell Science, for his vision, encouragement and support.

\section*{Author Contributions}

GRJ conceived, designed and implemented all experiments.
GRJ, RMD, and MMM wrote the paper.

\clearpage

\bibliographystyle{icml2013}


\clearpage

\appendix
\beginsupplement

\section{Supplementary Figures}

\begin{figure*}[htbp]
\centering
\includegraphics[width=0.9\linewidth]{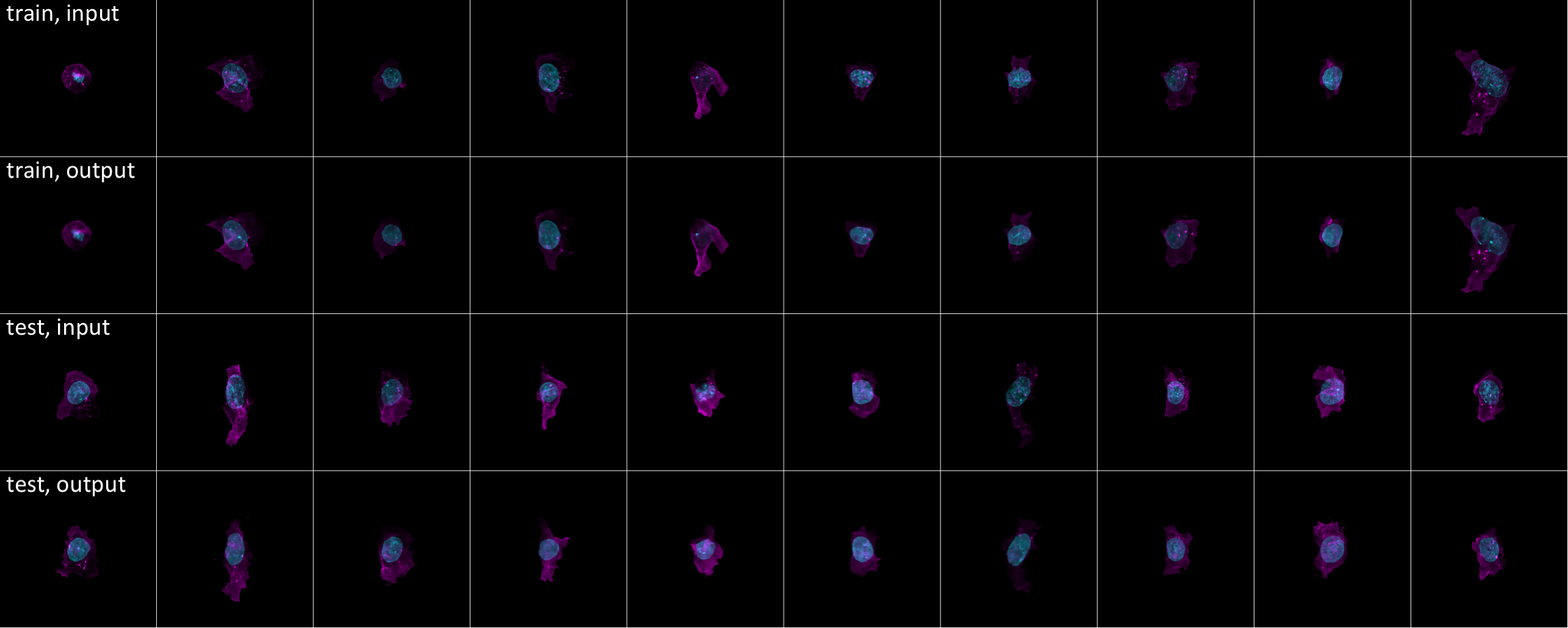}
\caption{Image input (rows 1 and 3) and reconstruction (rows 2 and 4) from the reference model, showing training set (above two rows), and test set (bottom two rows).}
\label{fig:traintest_ref}
\end{figure*}

\begin{figure*}[htbp]
\centering
\includegraphics[width=0.9\linewidth]{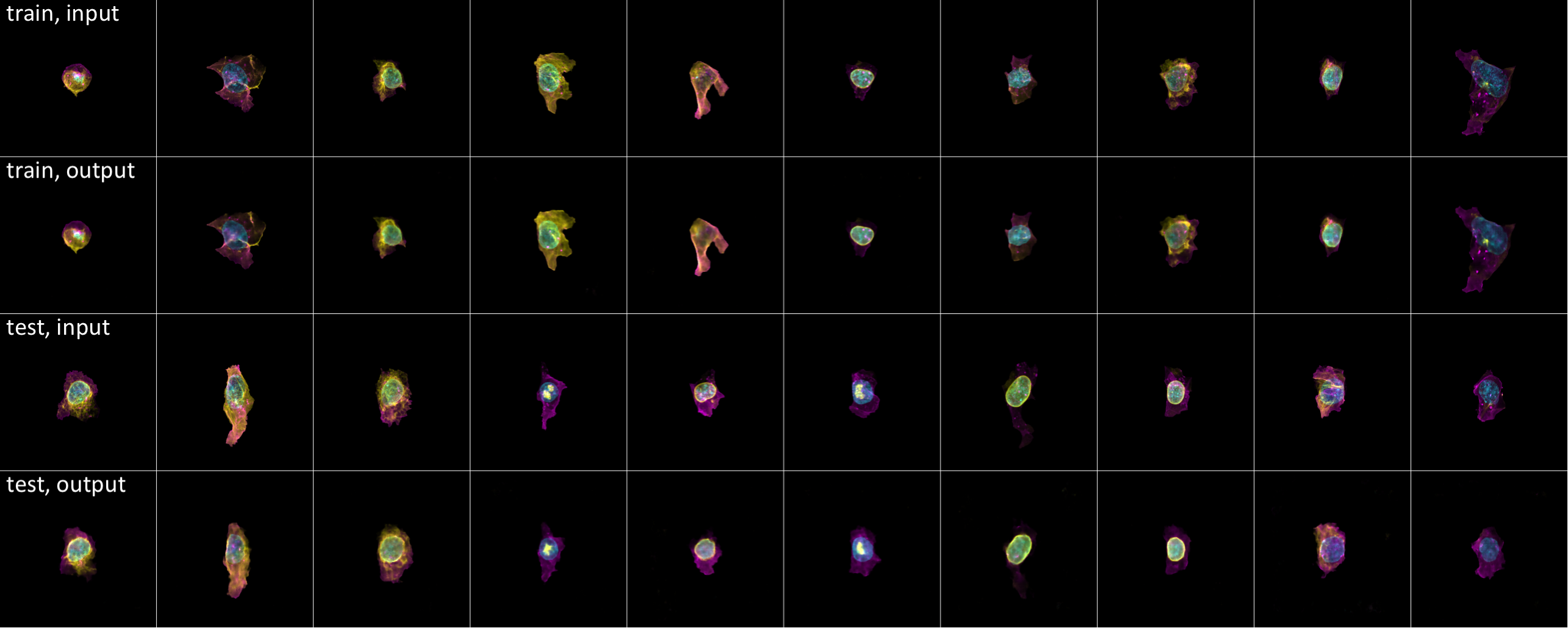}
\caption{Image input (rows 1 and 3) and reconstruction (rows 2 and 4) from the structure model, showing training set (above two rows), and test set (bottom two rows).}
\label{fig:traintest_struct}
\end{figure*}

\begin{figure*}[htbp]
  \centering
  \begin{subfigure}[b]{0.45\textwidth}
    \centering\includegraphics[width=.9\textwidth]{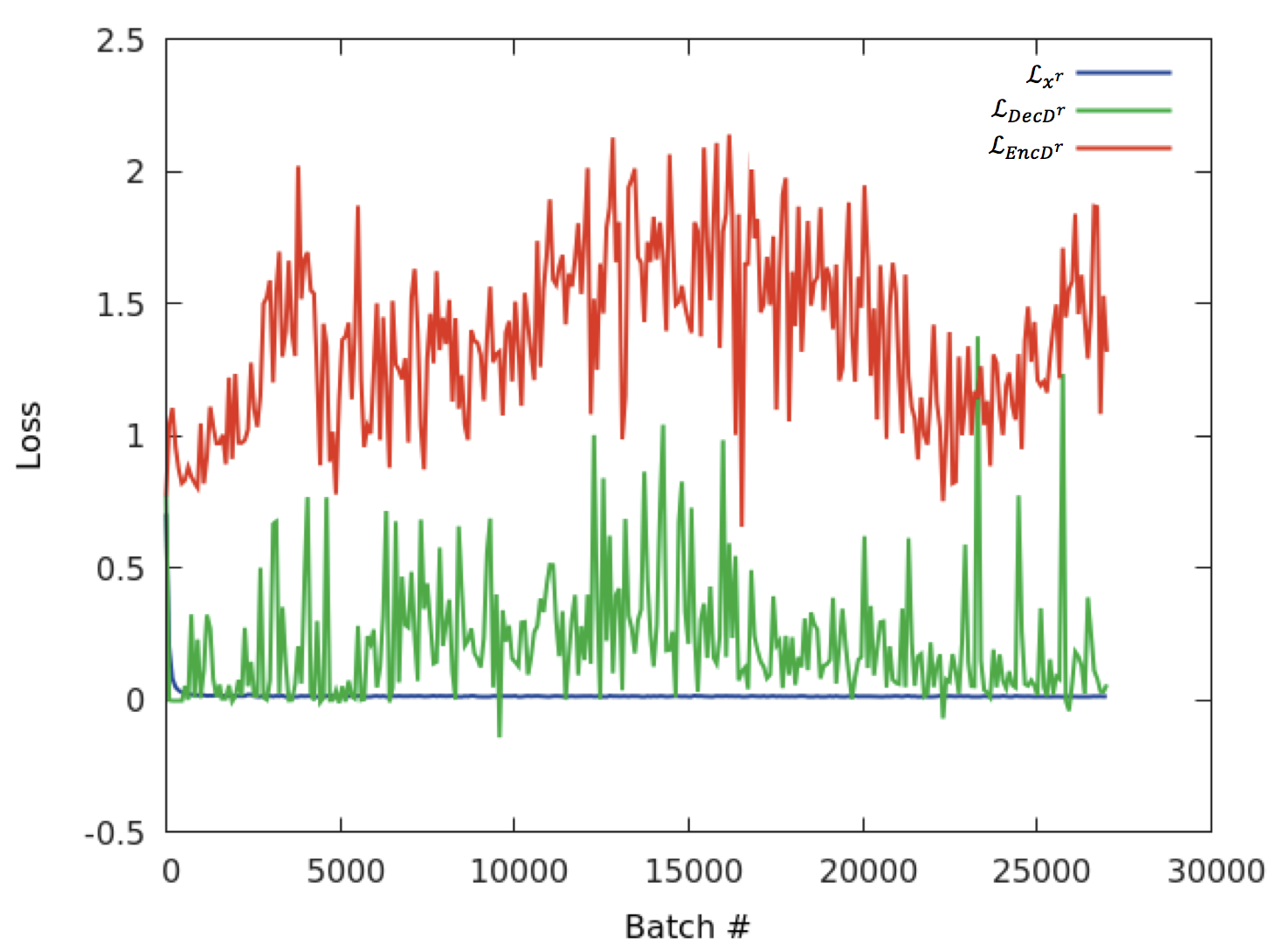}
    \caption{\label{fig:traincurves:fig1}}
  \end{subfigure}%
  \begin{subfigure}[b]{0.45\textwidth}
    \centering\includegraphics[width=.9\textwidth]{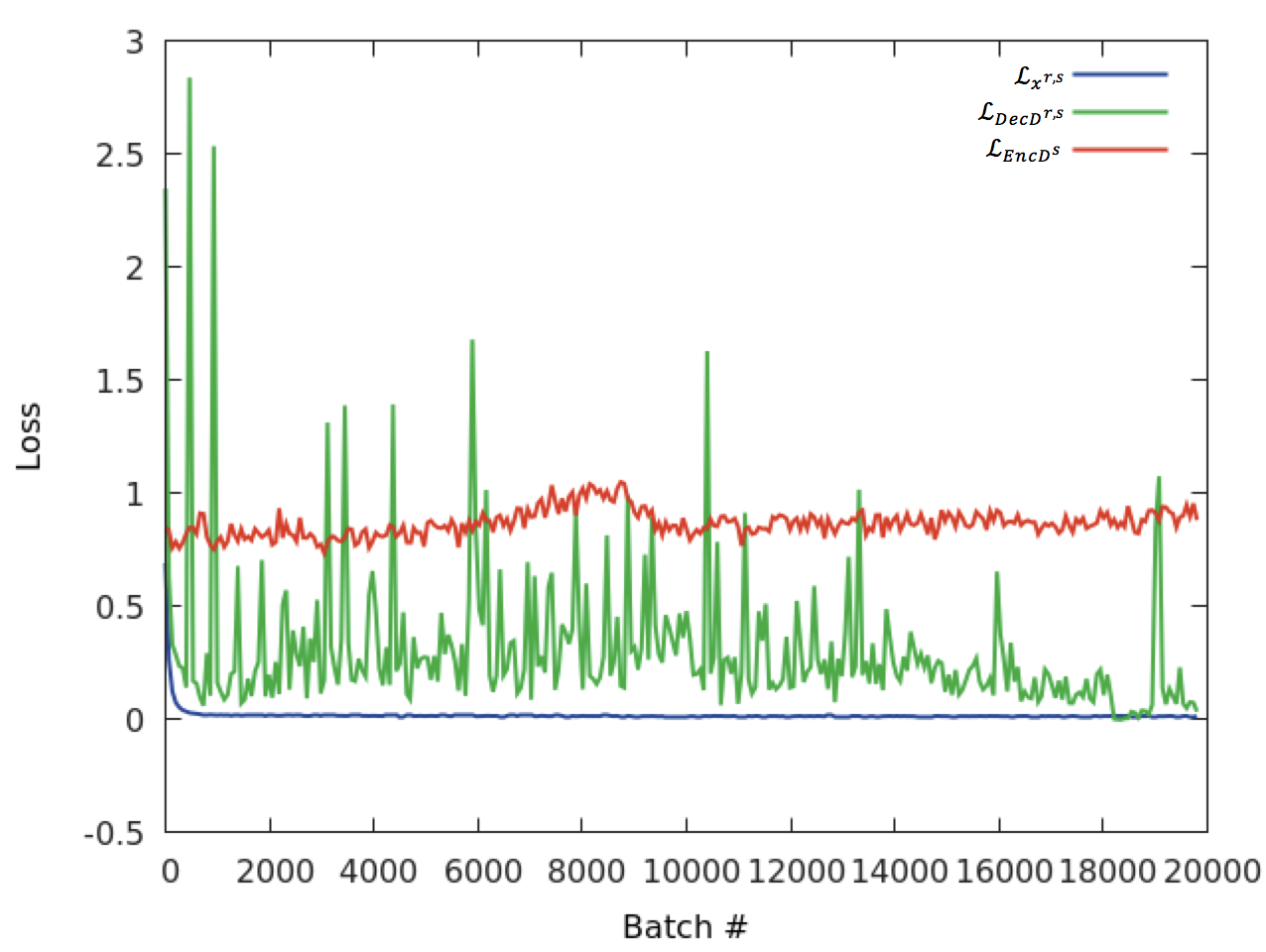}
    \caption{\label{fig:traincurves:fig2}}
  \end{subfigure}
  \caption{Training curves for the training of the reference model (\subref{fig:traincurves:fig1}) and conditional model (\subref{fig:traincurves:fig2})}
  \label{fig:traincurves}
\end{figure*}

\begin{figure*}[htbp]
  \centering
  \begin{subfigure}[b]{0.45\textwidth}
    \centering\includegraphics[width=.9\textwidth]{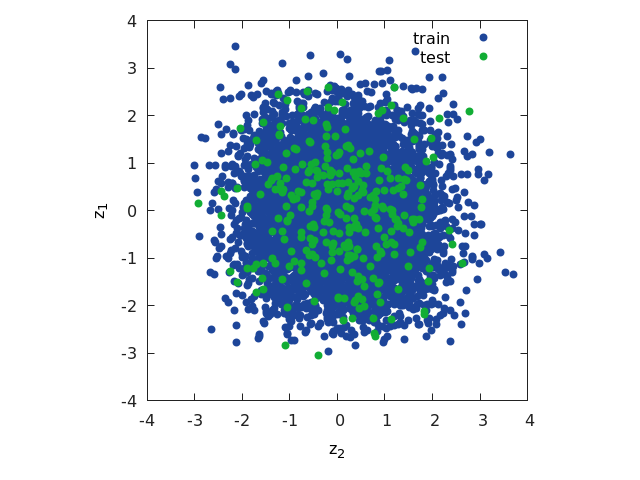}
    \caption{\label{fig:embed_ref:fig1}}
  \end{subfigure}%
  \begin{subfigure}[b]{0.45\textwidth}
    \centering\includegraphics[width=.9\textwidth]{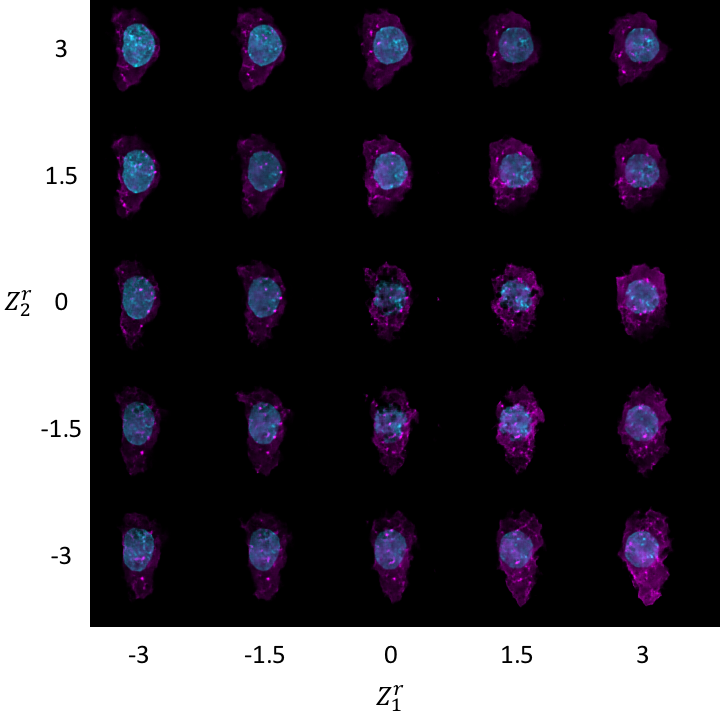}
    \caption{\label{fig:embed_ref:fig2}}
  \end{subfigure}
  \caption{(\subref{fig:fig1}) shows the first two dimensions of the reference structure latent space $\bm{Z}^r$.
  (~\subref{fig:traincurves:fig2}) shows the first two dimensions of the latent space sampleded at -3, -1.5, 0, 1.5 and 3 standard deviations in $\bm{Z}^r_1$ (horizontal) and $\bm{Z}^r_2$ (vertical).
  Images have been cropped for visualization purposes.}
  \label{fig:embed_ref}
\end{figure*}

\begin{figure*}[ht]
  \centering
  \begin{subfigure}[b]{0.45\textwidth}
    \centering\includegraphics[width=.9\textwidth]{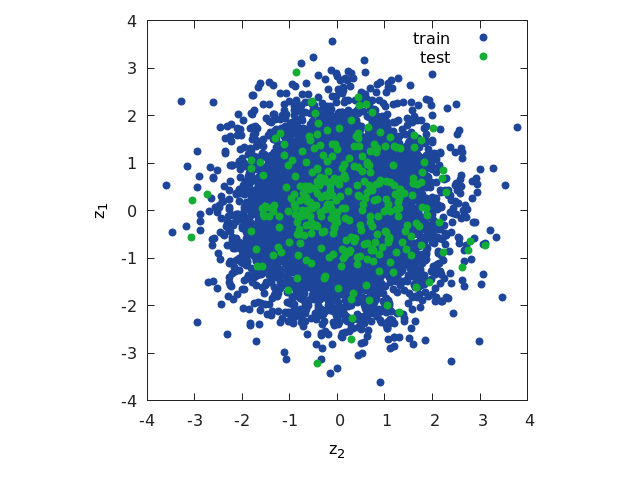}
    \caption{\label{fig:fig1}}
  \end{subfigure}%
  \begin{subfigure}[b]{0.45\textwidth}
    \centering\includegraphics[width=.9\textwidth]{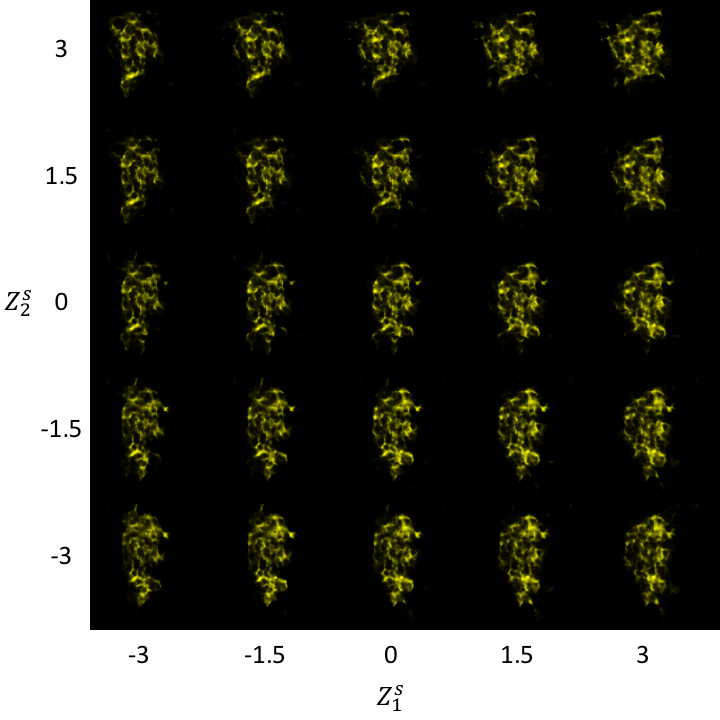}
    \caption{\label{fig:fig2}}
  \end{subfigure}
  \caption{(\subref{fig:fig1}) shows the first two dimensions of the reference structure latent space $\bm{Z}^s$.
  (~\subref{fig:traincurves:fig2}) shows the first two dimensions of the TOM20 latent space sampleded at -3, -1.5, 0, 1.5 and 3 standard deviations in $\bm{Z}^s_1$ (horizontal) and $\bm{Z}^s_2$ (vertical).
  Images have been cropped for visualization purposes.}
    \label{fig:embed_struct}
\end{figure*}

\begin{figure*}[ht]
  \centering
  \begin{subfigure}[b]{0.45\textwidth}
    \centering\includegraphics[width=.9\textwidth]{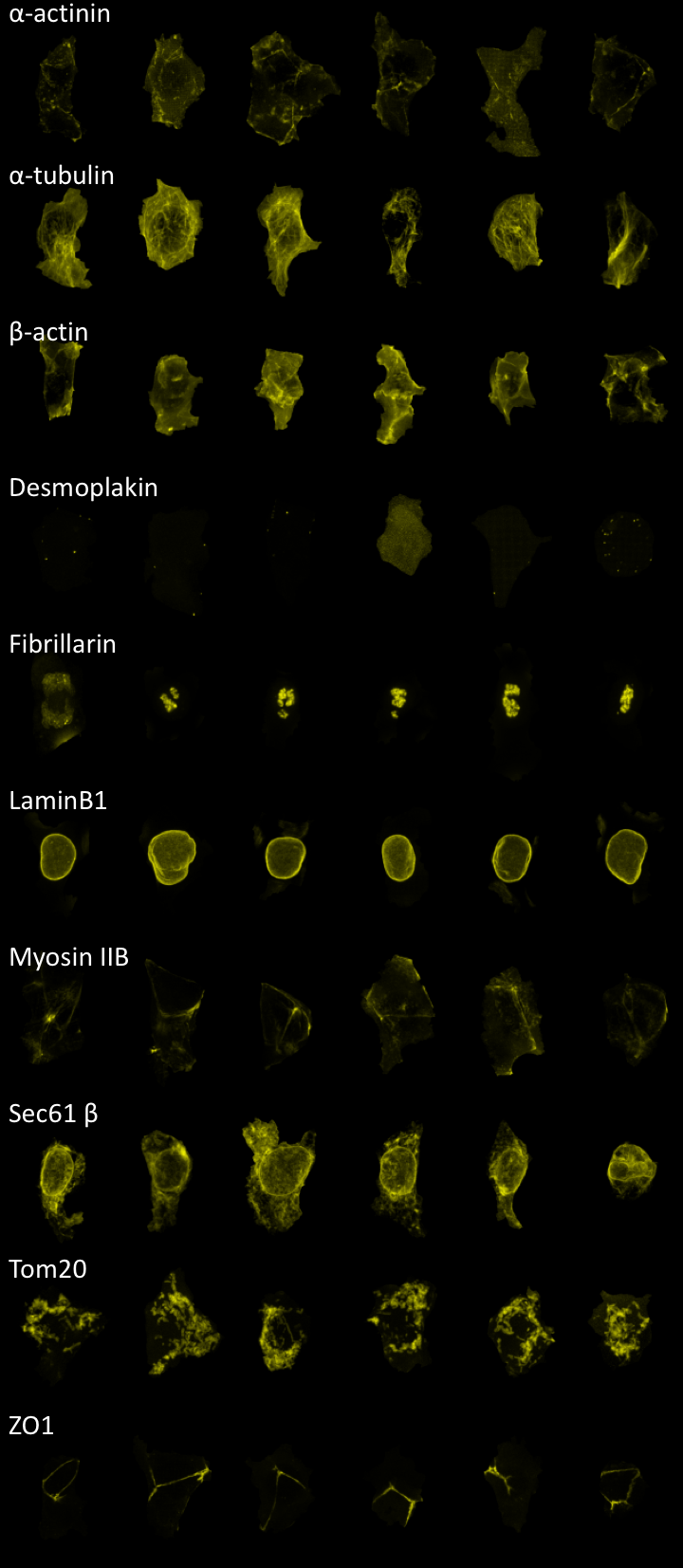}
    \caption{\label{fig:cells_real_struct_only}}
  \end{subfigure}%
  \begin{subfigure}[b]{0.45\textwidth}
    \centering\includegraphics[width=.9\textwidth]{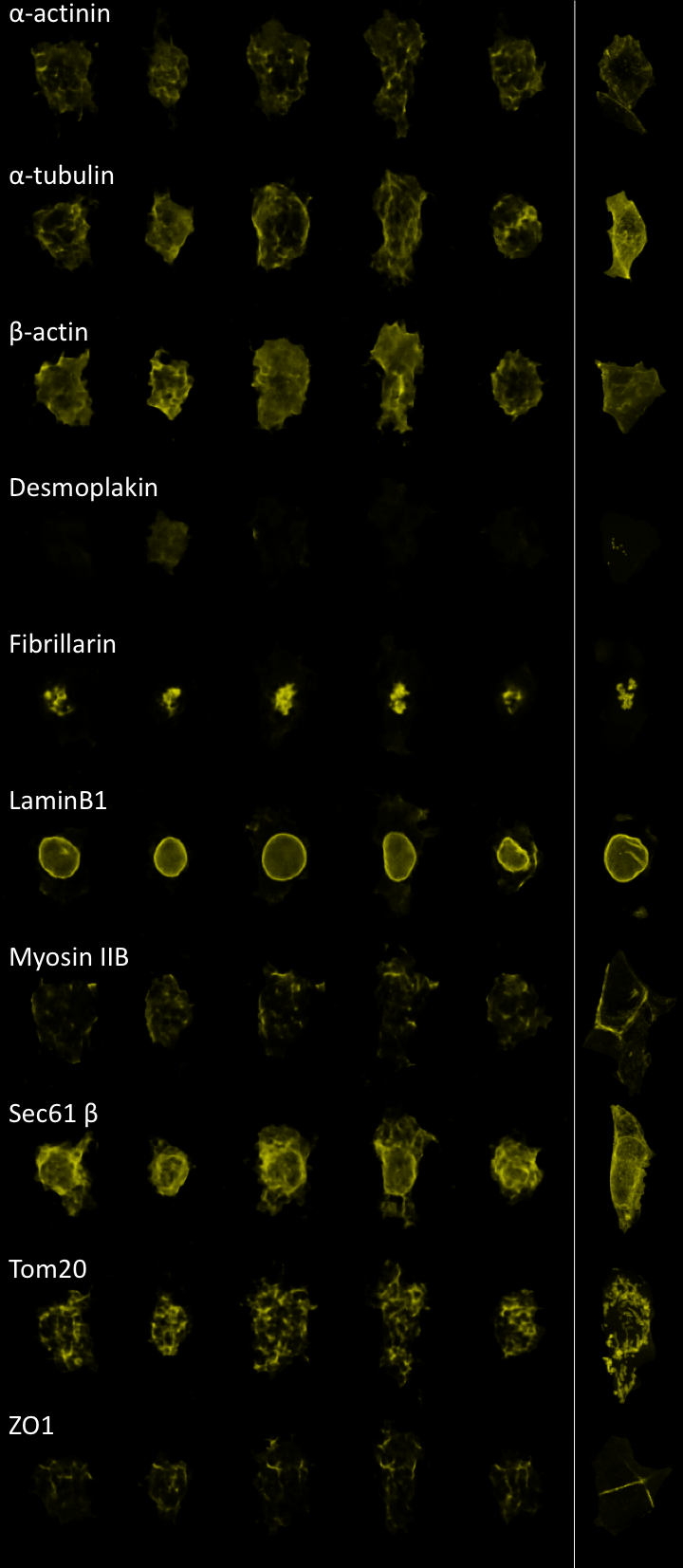}
    \caption{\label{fig:pred_struct_struct_only}}
  \end{subfigure}
  \caption{(\subref{fig:cells_real_struct_only}) Example structure channels for each of the 10 labeled structures in this paper and (\subref{fig:pred_struct_struct_only}) predicted most probable localization patterns for selected cells from each labeled pattern.
  The first 5 columns show the maximum likelihood localization for the corresponding structures given the the same cell and nuclear shape.
  The last column shows a observed cell with that labeled structure.
  Rows correspond to structure types.
  Images have been cropped for visualization purposes.}
\label{fig:struct_only}
\end{figure*}

\clearpage

\section{Model Architectures}
\label{sec:architecture}

\begin{table}[htbp]
\centering \textsc{ \small{
\begin{tabular}{lcc}
\hline
\abovespace
$4\times4$ 64   conv $\downarrow$ & BNorm & PReLU \\
$4\times4$ 128  conv $\downarrow$ & BNorm & PReLU \\
$4\times4$ 256  conv $\downarrow$ & BNorm & PReLU \\
$4\times4$ 512  conv $\downarrow$ & BNorm & PReLU \\
$4\times4$ 1024 conv $\downarrow$ & BNorm & PReLU \\
$4\times4$ 1024 conv $\downarrow$ & BNorm & PReLU \\
\belowspace
$\abs{Z^r}$ FC                    & BNorm & \\
\hline
\end{tabular}
}}
\vskip -0.1in
\caption{Architecture of $\Enc^r$}
\label{tbl:arch_Enc_r}
\end{table}

\begin{table}[htbp]
\centering \textsc{ \small{
\begin{tabular}{lcc}
\hline
\abovespace
1024 FC & BNorm & PReLU \\
$4\times4$ 1024 conv $\uparrow$ & BNorm & PReLU \\
$4\times4$ 512 conv $\uparrow$ & BNorm & PReLU \\
$4\times4$ 256  conv $\uparrow$ & BNorm & PReLU \\
$4\times4$ 128  conv $\uparrow$ & BNorm & PReLU \\
$4\times4$ 64  conv $\uparrow$ & BNorm & PReLU \\
\belowspace
$4\times4$ $\abs{r}$ conv $\uparrow$ & BNorm & sigmoid \\
\hline
\end{tabular}
}}
\vskip -0.1in
\caption{Architecture of $\Dec^r$}
\label{tbl:arch_Dec_r}
\end{table}

\begin{table}[htbp]
\centering \textsc{ \small{
\begin{tabular}{lcc}
\hline
\abovespace
1024 FC &       & Leaky RelU \\
1024 FC & BNorm & Leaky RelU \\
 512 FC & BNorm & Leaky RelU \\
\belowspace
   1 FC &       & Sigmoid \\
\hline
\end{tabular}
}}
\vskip -0.1in
\caption{Architecture of $\EncD^r$ and $\EncD^s$}
\label{tbl:arch_EncD}
\end{table}

\begin{table}[htbp]
\centering \textsc{ \small{
\begin{tabular}{lcc}
\hline
\abovespace
$+$ White Noise $\sigma = 0.05$ \\
$4\times4$  64 conv $\downarrow$ & BNorm & LeakyReLU \\
$4\times4$ 128 conv $\downarrow$ & BNorm & LeakyReLU \\
$4\times4$ 256 conv $\downarrow$ & BNorm & LeakyReLU \\
$4\times4$ 512 conv $\downarrow$ & BNorm & LeakyReLU \\
$4\times4$ 512 conv $\downarrow$ & BNorm & LeakyReLU \\
\belowspace
$4\times4$   1 conv $\downarrow$ &       & sigmoid \\
\hline
\end{tabular}
}}
\vskip -0.1in
\caption{Architecture of $\DecD^r$}
\label{tbl:arch_DecD_r}
\end{table}

\begin{table}[htbp]
\centering \textsc{ \small{
\begin{tabular}{lcc}
\hline
\abovespace
$4\times4$   64 conv $\downarrow$ & BNorm & PReLU \\
$4\times4$  128 conv $\downarrow$ & BNorm & PReLU \\
$4\times4$  256 conv $\downarrow$ & BNorm & PReLU \\
$4\times4$  512 conv $\downarrow$ & BNorm & PReLU \\
$4\times4$ 1024 conv $\downarrow$ & BNorm & PReLU \\
$4\times4$ 1024 conv $\downarrow$ & BNorm & PReLU \\
\belowspace
\{K FC, $\abs{Z^r}$ FC, $\abs{Z^s}$ FC\} & \{BNorm, BNorm, BNorm\} & \{Softmax, \phantom{N}, \phantom{N}\}\\
\hline
\end{tabular}
}}
\vskip -0.1in
\caption{Architecture of $\Enc^{r,s}$}
\label{tbl:arch_Enc_rs}
\end{table}

\begin{table}[htbp]
\centering \textsc{ \small{
\begin{tabular}{lcc}
\hline
\abovespace
1024 FC & BNorm & PReLU \\
$4\times4$ 1024 conv $\uparrow$ & BNorm & PReLU \\
$4\times4$ 512 conv $\uparrow$ & BNorm & PReLU \\
$4\times4$ 256  conv $\uparrow$ & BNorm & PReLU \\
$4\times4$ 128  conv $\uparrow$ & BNorm & PReLU \\
$4\times4$ 64  conv $\uparrow$ & BNorm & PReLU \\
\belowspace
$4\times4$ $\abs{r+s}$ conv $\uparrow$   & BNorm & sigmoid \\
\hline
\end{tabular}
}}
\vskip -0.1in
\caption{Architecture of $\Dec^{r,s}$}
\label{tbl:arch_Dec_rs}
\end{table}

\begin{table}[htbp]
\centering \textsc{ \small{
\begin{tabular}{lcc}
\hline
\abovespace
$+$ White Noise $\sigma = 0.05$ \\
$4\times4$  64 conv $\downarrow$ & BNorm & LeakyReLU \\
$4\times4$ 128 conv $\downarrow$ & BNorm & LeakyReLU \\
$4\times4$ 256 conv $\downarrow$ & BNorm & LeakyReLU \\
$4\times4$ 512 conv $\downarrow$ & BNorm & LeakyReLU \\
$4\times4$ 512 conv $\downarrow$ & BNorm & LeakyReLU \\
\belowspace
$4\times4$ K+1 conv $\downarrow$ &       & sigmoid \\
\hline
\end{tabular}
}}
\vskip -0.1in
\caption{Architecture of $\DecD^{r,s}$}
\label{tbl:arch_DecD_rs}
\end{table}

\clearpage

\section{Data}

\begin{table}[htbp]
\centering
\begin{tabular}{lrrr}
\hline
\abovespace\belowspace
Labeled Structure & \#total & \#train & \#test\\
\hline
\abovespace
\textalpha-actinin  &  493 &  462 & 31 \\
\textalpha-tubulin  & 1043 & 1002 & 41 \\
\textbeta-actin     &  542 &  513 & 29 \\
Desmoplakin         &  229 &  219 & 10 \\
Fibrillarin         &  988 &  953 & 35 \\
Lamin B1            &  785 &  739 & 46 \\
Myosin IIB          &  157 &  149 &  8 \\
Sec61\textbeta     &  835 &  784 & 51 \\
TOM20               &  771 &  723 & 48 \\
\belowspace
ZO1                 &  234 &  229 &  5 \\
\hline
\end{tabular}
\vskip -0.1in
\caption{Labeled structures and their train/test split}
\label{tbl:traintest}
\end{table}

\begin{table*}[htbp]
\centering
\begin{tabular}{rcccccccccc}
\textalpha-actinin  & 22 & 2 & 5 & 1 & 0 & 0 & 0 & 0 & 0 & 1 \\
\textalpha-tubulin  & 0 & 36 & 3 & 0 & 0 & 0 & 0 & 1 & 1 & 0 \\
\textbeta-actin     & 3 & 7 & 19 & 0 & 0 & 0 & 0 & 0 & 0 & 0 \\
Desmoplakin         & 1 & 0 & 1 & 7 & 0 & 0 & 0 & 0 & 0 & 1 \\
Fibrillarin         & 0 & 0 & 0 & 0 & 35 & 0 & 0 & 0 & 0 & 0 \\
Lamin B1            & 0 & 0 & 0 & 0 & 0 & 46 & 0 & 0 & 0 & 0 \\
Myosin IIB          & 2 & 0 & 0 & 1 & 0 & 0 & 0 & 0 & 1 & 4 \\
Sec61\textbeta     & 0 & 1 & 0 & 0 & 0 & 0 & 0 & 50 & 0 & 0 \\
TOM20               & 0 & 1 & 0 & 0 & 0 & 0 & 0 & 0 & 47 & 0 \\
ZO1                 & 1 & 0 & 0 & 1 & 0 & 0 & 2 & 0 & 0 & 1 \\
\end{tabular}
\vskip -0.1in
\caption{Labeled structure class prediction results on hold out}
\label{tbl:confmat}
\end{table*}

\end{document}